\pdfoutput=1
\documentclass[10pt,twocolumn,letterpaper]{article}
\usepackage[pagenumbers]{cvpr} %

\usepackage{enumitem}
\usepackage{graphicx}
\usepackage{amsmath}
\usepackage{amssymb}
\usepackage{booktabs}
\usepackage{multirow}
\usepackage{xcolor}
\usepackage{amsmath,amsfonts,bm}
\usepackage{adjustbox}
\usepackage{floatrow}
\newfloatcommand{capttabbox}{table}[][\FBwidth]
\definecolor{supconPurple}{RGB}{153, 51, 255}
\usepackage[pagebackref,breaklinks,colorlinks]{hyperref}

\usepackage[capitalize]{cleveref}
\crefname{section}{Sec.}{Secs.}
\Crefname{section}{Section}{Sections}
\Crefname{table}{Table}{Tables}
\crefname{table}{Tab.}{Tabs.}
\newcommand{\printfnsymbol}[1]{%
  \textsuperscript{\@fnsymbol{#1}}%
}

\def\cvprPaperID{7076} %
\def\confName{CVPR}
\def\confYear{2022}

\begin{document}

\title{Probing Representation Forgetting in Supervised and Unsupervised Continual Learning}
\author{\parbox{\textwidth}{\centering
    MohammadReza Davari$^{1,2}\;$\thanks{Equal contribution, name order randomized.\\This research was partially funded by NSERC Discovery Grant RGPIN-2021-04104 and  RGPIN-2019-05729. We acknowledge resources provided by Compute Canada and Calcul Quebec. Correspondence to: \texttt{\\m\_davari@live.concordia.ca, nader.asadi@concordia.ca, \\eugene.belilovsky@concordia.ca}} \hspace{-10pt}
    \qquad Nader Asadi$^{1,2 \; *}$ \hspace{-10pt}
    \qquad Sudhir Mudur$^{1}$ \hspace{-10pt}\\
    \qquad Rahaf Aljundi $^{3}$\hspace{-10pt}
    \qquad Eugene Belilovsky$^{1,2}$} \vspace{5pt}\\
$^1$ Concordia University~ $^2$ Mila -- Quebec AI Institute ~ $^3$ Toyota Motor Europe \\
}

\if 0
\author{MohammadReza Davari
\thanks{Equal contribution, name order randomized.\\This research was partially funded by NSERC Discovery Grant RGPIN-2021-04104 and  RGPIN-2019-05729. We acknowledge resources provided by Compute Canada and Calcul Quebec. }\\
Concordia University and Mila\\
{\tt\small m\_davari@live.concordia.ca}
\and
Nader Asadi$^{*}$\\
Concordia University and Mila\\
{\tt\small nader.asadi@concordia.ca}
\and
Sudhir Mudur\\
Concordia University\\
{\tt\small sudhir.mudur@concordia.ca}
\and
Rahaf Aljundi\\
Toyota Motor Europe\\
{\tt\small rahaf.al.jundi@toyota-europe.com}
\and
Eugene Belilovsky\\
Concordia University and Mila\\
{\tt\small eugene.belilovsky@concordia.ca}
}

\fi
\maketitle

\begin{abstract}
\vspace{-9pt}
Continual Learning (CL) research typically focuses on tackling the phenomenon of catastrophic forgetting in neural networks. Catastrophic forgetting is associated with an abrupt loss of knowledge previously learned by a model when the task, or more broadly the data distribution, being trained on changes. In supervised learning problems this forgetting, resulting from a change in the model's representation, is typically measured or observed by evaluating the decrease in old task performance. However, a model’s representation can change without losing knowledge about prior tasks. In this work we consider the concept of representation forgetting, observed by using the difference in performance of an optimal linear classifier before and after a new task is introduced. Using this tool we revisit a number of standard continual learning benchmarks and observe that, through this lens, model representations trained without any explicit control for forgetting often experience small representation forgetting and can sometimes be comparable to methods which explicitly control for forgetting, especially in longer task sequences. We also show that representation forgetting can lead to new insights on the effect of model capacity and loss function used in continual learning. Based on our results, we show that a simple yet competitive approach is to learn representations continually with standard supervised contrastive learning while constructing prototypes of class samples when queried on old samples. 

\vspace{-18pt}
\end{abstract}

\section{Introduction}
\vspace{-5pt}
Continual Learning (CL) is concerned with methods for learners to manage changing training data distributions. The goal is to acquire new knowledge from new data distributions while not forgetting previous knowledge. A common scenario is CL in the classification setting, where the class labels presented to the learner change over time. In this scenario, a phenomenon known as catastrophic forgetting has been observed~\cite{mccloskey1989catastrophic,goodfellow2013empirical}. This phenomenon is often described as a loss of knowledge about previously seen data and is observed in the classification setting as a decrease in accuracy. 

Deep learning has been traditionally motivated as an approach, which can automatically learn representations~\cite{bengio2013representation}, forgoing the need to design handcrafted features. Indeed representation learning is at the core of deep learning methods in supervised and unsupervised settings~\cite{Goodfellow-et-al-2016-Book}. In the case of many practical scenarios we may not simply be interested in the final performance of the model, but also the usefulness of the learned features for various downstream tasks~\cite{ILSVRC15}. Although a model's representation may change, sometimes drastically at task boundaries~\cite{caccia2021reducing}, this does not necessarily imply a loss of useful information and may instead correspond to a simple transformation. For example, consider a standard multi-head CL setting, where each task shares a representation and only differs through task specific ``heads". A permutation of the features leading into the classification heads leads to total catastrophic forgetting as measured by standard approaches as the task heads no longer match with the representations. However, this does not correspond to a loss of knowledge about the data, nor a less useful representation. Indeed recent works have highlighted the importance of fast remembering versus catastrophic forgetting~\cite{he2019task, hadsell2020embracing}, a looser continual learning requirement where in the task performance may decrease but the agent is able to recover rapidly upon observing a few samples from the previous task. In this light, maintaining a useful representation, which facilitates rapid recovery, is as important as maintaining high performance for the task.

CL envisions having learners operate over long time horizons while continually maintaining old knowledge and integrating new knowledge. Hence, in addition to directly measuring the performance on previous tasks using the last layer classifiers, it is sensible to consider the usefulness of their representations for previous tasks.  In this paper we highlight that traditional approaches of evaluating forgetting are unable to properly disambiguate trivial changes in the features (e.g. permutation) from abrupt losses of useful representations. We instead use optimal Linear Probes (LP), commonly used to study unsupervised representations~\cite{chen2020simple} and intermediate layer representations~\cite{oyallon2017building,zeiler2014visualizing}, to evaluate CL algorithms and their effectiveness.

We revisit several CL settings and benchmarks and measure forgetting using LP. Our focus is particularly on re-evaluating finetuning approaches that do not apply explicit control for the non-iid nature of continual learning. We observe that in many commonly studied cases of catastrophic forgetting, the representations under naive finetuning approaches, undergo minimal forgetting, without losing critical task information. 

Our major contributions in this work are as follows. 
First we bring three new significant insights obtained and demonstrated through extensive experimental analysis:
\vspace{-0.2em}
\begin{enumerate}[labelindent=3pt]
\setlength\itemsep{-0.15em}
\item In a number of CL settings the observed accuracy can be a misleading metric for studying forgetting, particularly when compared to finetuning approaches 
\item Naive training with SupCon~\cite{NEURIPS2020_d89a66c7} or SimCLR (in the unsupervised case) have advantageous properties for continual learning, particularly in longer sequences. 
\item By using LP based evaluation, forgetting is clearly decreased for wider and deeper models, which is not seen that clearly from earlier observed accuracy.
\end{enumerate}
\vspace{-0.2em}
Secondly, we suggest a simple approach to facilitate fast remembering, which does not require using a large memory during training; it relies only on a small memory combined with SupCon based finetuning.

\vspace{-4pt}
\section{Related Work} \vspace{-4pt}
The design of CL methods is often focused on mitigating the catastrophic forgetting phenomenon, with aspects such as maximization of forward and backward transfer between tasks taken as secondary~\cite{lopez2017gradient}. One class of methods focuses on bypassing this problem by growing architectures over time as new tasks arrive~\cite{aljundi2016expert,li2019learn,rosenfeld2018incremental, rusu2016progressive}. Under the fixed architecture setting, one can identify two main categories. The first category of methods rely on storing and re-using samples from the previous history while learning new ones; this includes approaches such as GEM~\cite{lopez2017gradient} and ER~\cite{chaudhry2019continual}. The second category of methods encode the knowledge of the previous tasks in a prior that is used to regularize the training of the new task ~\cite{li2017learning,kirkpatrick2016overcoming,Zenke2017improved,aljundi2018memory,nguyen2017variational}. A classic method in this vein is Learning without Forgetting (LwF)~\cite{li2017learning}, which mitigates forgetting by a regularization term that distills knowledge~\cite{hinton2015distilling} from the earlier tasks. The network representations from earlier stages are recorded, and are used during training for a new task to regularize the objective by distilling knowledge from the earlier state of the network. 
Similarly, Elastic Weight Consolidation (EWC)~\cite{huszar2017quadratic} preserves the knowledge of the past tasks through a quadratic penalty on the network parameters important to the earlier tasks. The importance of the parameters is approximated via the diagonal of the Fisher information matrix~\cite{myung2003tutorial}. The scale of the importance matrix, $\lambda$, determines the network's preference towards preserving old representations or acquiring new ones for the current task.    
In~\cref{sec:observed-vs-lp-acc} we examine the effectiveness of these approaches in mitigating representation forgetting.

Recent works on elucidating the nature of catastrophic forgetting have examined the influence of task sequence~\cite{nguyen2019toward}, network architecture~\cite{arora-etal-2019-lstm}, and change in representation similarity~\cite{ramasesh2020anatomy}. Our work is related in spirit to~\cite{ramasesh2020anatomy} as we pursue measuring how much forgetting has occurred on the learned representation and we additionally study this for intermediate representations. One significant difference is that in~\cite{ramasesh2020anatomy}, the authors use linear CKA (centred Kernel Alignment)~\cite{kornblith2019similarity} to measure the similarity between intermediate representations influenced by forgetting, while in our work we measure how much forgetting has occurred on the representations using LP.

Several recent works have also studied the behavior of networks with increasing model capacity. In \cite{anonymous2022effect} the authors examine several common architectures under the task incremental setting and demonstrate that pre-training is essential to combat forgetting and to achieve high performance on all tasks. They conclude that training only with larger models yields no benefit for continual learning. Our analysis revisit this setting and take a closer look at how representation forgetting is affected with increase in model width and depth. 

Several works~\cite{rebuffi2017icarl} have focused on modifying the last layer of a classification network to make more effective use of the representation for prior tasks. This indirectly highlights the fact that the last layer can be modified to yield better performance on prior tasks. Particularly~\cite{rebuffi2017icarl,mai2021supervised} use a buffer of old examples at training time to improve learning and then use them at evaluation time to construct a class mean prototype. This allows for more effective use of the representations of the network. These works consider settings where the CL methods are used to control forgetting, while we also emphasize that naive continuation of training under task shift can yield strong representations. Our work can also be seen as both a way to explain and to motivate the need for such approaches.  

Self-supervised learning (SSL) is becoming increasingly popular in visual representation learning. Some of the best performing methods rely on contrastive learning~\cite{chen2020simple,he2019moco}. These methods have been recently evaluated in a limited continual learning setting~\cite{hu2021well} where a sequence was trained on non-iid unsupervised streaming data and then applied in transfer learning settings on multiple datasets. However, forgetting was not evaluated. In contrast our work, which also uses a SSL loss, focuses on the LP evaluation and the study of representation forgetting with respect to previously seen distributions. Contrastive methods are also often used in the supervised setting, for example, the recently popular SupCon loss~\cite{NEURIPS2020_d89a66c7}.
In~\cite{caccia2021reducing} and~\cite{mai2021supervised} the use of SupCon is proposed in the online class-incremental setting in combination with experience replay. Our work too considers SupCon as one of the supervised representation learning approaches. However distinct from the other works we consider it in the offline task-incremental setting. We do not look at its use in combination with replay or other approaches, but study the effect of standard finetuning with SupCon loss, distinct from~\cite{caccia2021reducing}, where it is used to facilitate separation of contrasts between old and new classes, specific to the class incremental setting.

\section{Background and Methods}
In this section we review the key tools used in our analysis including linear probes, centered kernel alignment, and contrastive loss functions. Finally we discuss how the nearest mean of exemplars approach can be used in the context of non-rehearsal based methods, such as finetuning with SupCon, as a simple continual learning method that also facilitates rapid remembering. 
\subsection{Linear Probes for Representation Forgetting}
\vspace{-3pt}
Following the work in SSL~\cite{chen2020simple} and in the analysis of intermediate representations~\cite{zeiler2014visualizing} we evaluate the adequacy of representations by an optimal linear classifier using training data from the original task. A linear classifier is trained on top of the frozen activations of the base network given the training instances of a certain dataset. The test set accuracy obtained by using LP on that dataset is used as a proxy to measure the quality of the representations. The difference in performance of the LP before and after a new task is introduced, acts as a surrogate measure to the amount of forgetting observed by the representations and is referred to as representation forgetting.

Formally, for a given model $f_{\theta_i}$ computed from time step $i$ of a task sequence, we compute its optimal classifier $W_i^*=\arg \min_{W_i} \mathcal{L}(W_if_{\theta_i}(X_i),Y_i)$, where $\mathcal{L}$ is the objective function, and $X_i$ and $Y_i$ correspond to the data from task $i$.
To assess representation forgetting between $\theta_a$ and a model at a later point in the sequence, say $\theta_b$, we evaluate $T(W_af_{\theta_a}(X_a),Y_a)-T(W_bf_{\theta_b}(X_a),Y_a)$ where $T$ is the task performance (\eg accuracy). 

\subsection{Centered Kernel Alignment}
\vspace{-3pt}
CKA~\cite{kornblith2019similarity} is a recent popular approach to compare representations. It has been commonly used to compare representations across depth as well as across models from different tasks in the CL settings~\cite{ramasesh2020anatomy}.
Given a dataset comprised of $m$ samples, and their representations $X$ and $Y$, with features $n_x$ and $n_y$ respectively, \ie $X \in \mathbb{R}^{m \times n_x}$ and $Y \in \mathbb{R}^{m \times n_y}$, the, typically used, linear CKA between $X$ and $Y$ is given as $\frac{\left\| Y^TX \right\|^2_F}{\left\|X^TX\right\|^2_F\left\|Y^TY\right\|^2_F}$. This similarity metric has the advantage of being invariant to scaling and orthogonal transformations. However, being a simple linear alignment comparison it is not invariant to general classes of invertible transformations. Furthermore, relative comparisons of CKA metrics are challenging to interpret compared to task performance degradation. CKA has been used in~\cite{ramasesh2020anatomy} to compare the intermediate representations of a model in consecutive task increments $t$ and $t+1$. Ramasesh \etal~\cite{ramasesh2020anatomy} proxy the CKA similarity between the intermediate representations of the model $f_{\theta_{t}}$ and $f_{\theta_{t+1}}$ to measure representation forgetting. Thus, under this paradigm, a high value of CKA similarity is interpreted as minimal representation forgetting. One limitation of this metric for representation forgetting is its inability to distinguish between positive and negative backward transfer (see \cref{sec:depthwise-probe-and-comparison-to-cka}). This is addressed when measuring representation forgetting via LP.

\newcommand{\simm}{\texttt{sim}} 
\newcommand{\ff}{f_{\theta}}

\subsection{Supervised and Unsupervised Contrastive Loss}
Contrastive loss functions have become increasingly popular in representation learning, particularly visual representation learning. They have led to large advances in unsupervised learning~\cite{he2019moco,chen2020simple}. As well they are becoming a popular alternative to cross-entropy (CE) in the supervised setting~\cite{NEURIPS2020_d89a66c7, graf2021dissecting}, referred to as SupCon. Given a representation $f_{\theta}$, often consisting of a primary network and a projection, the SupCon loss for a minibatch $X$ is given by:
\begin{equation*}
\sum_{\mathbf{x}_i \in \mathbf{X}}\frac{-1}{|P(i)|}\sum_{\mathbf{x}_p \in P(i)}\log\frac{\simm\big(\ff(\mathbf{x}_p), \ff(\mathbf{x}_i)\big)}{\sum_{\mathbf{x}_a \in \mathbf{X}/x_i}
\simm \big(\ff(\mathbf{x}_a), \ff(\mathbf{x}_i)\big)}
\end{equation*}
Where $\simm(a,b) = \exp (\frac{a^Tb}{\tau\|a\|\|b\|})$ and $P(i)$ represents the same class samples as $x_i$ from the minibatch, and the denominator is taken over all other samples. Note that we consider SupCon in the naive setting, thus all minibatches are from the current task in our evaluations of SupCon. Similar to this, in the unsupervised setting the SimCLR loss~\cite{chen2020simple} is given by:
\setlength{\belowdisplayskip}{2pt} \setlength{\belowdisplayshortskip}{2pt}
\setlength{\abovedisplayskip}{2pt} \setlength{\abovedisplayshortskip}{2pt}
\begin{equation*}
-\sum_{\mathbf{x}_i \in \mathbf{X}}\log\frac{\simm\big(\ff(\mathbf{x}_p(i)), \ff(\mathbf{x}_i)\big)}{\sum_{\mathbf{x}_n \in \mathbf{A}(i)}
\simm \big(\ff(\mathbf{x}_n), \ff(\mathbf{x}_i)\big)}
\end{equation*}
Where $A(i)$ corresponds to all minibatch examples and their data augmentations except $x_i$, and $x_p(i)$ represents an augmented version of $x_i$.

\thisfloatsetup{floatrowsep=quad}
\begin{figure*}
\vspace{-1em}
\CenterFloatBoxes
\begin{floatrow}
\ffigbox[0.5\textwidth]{%
    \includegraphics[width=\linewidth]{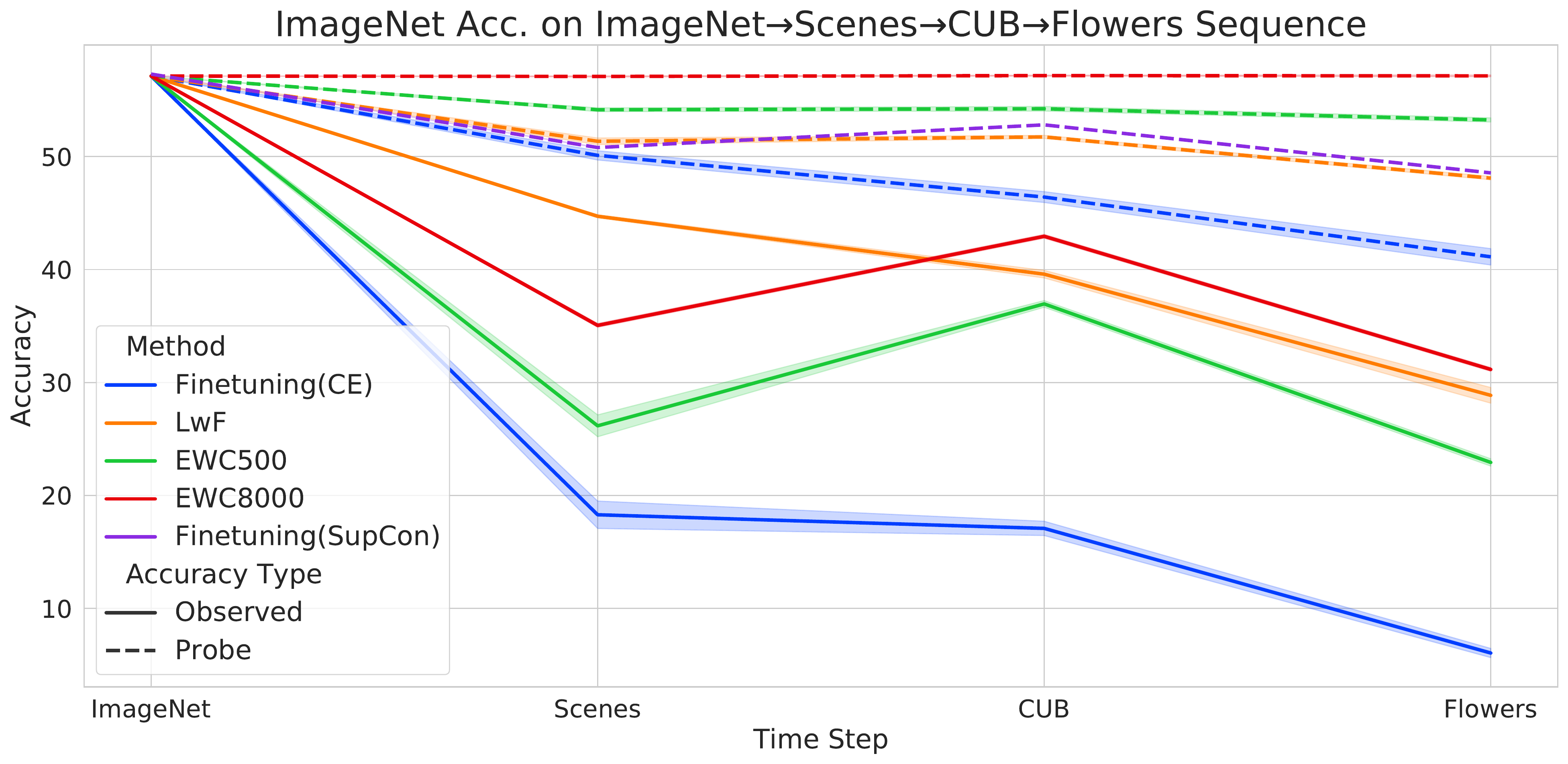}
}{%
  \caption{\small Performance on ImageNet during the sequence (ImageNet→Scenes→CUB→Flowers) using ResNet18. We observe that although observed accuracy heavily degrades the LP accuracy, in finetuning does not decay as drastically and can rival LP accuracy of methods such as LwF and EWC. Moreover, we observe that the LP Accuracy of SupCon training, which has no control for forgetting, outperforms the LwF, a method designed for CL. Note EWC with $\lambda=8k$ is the best performing method in terms of LP and observed acc., however it does not perform well on the current task (see \cref{tab:imagenet-scenes-cub}).
  \label{fig:imagenet-scenes-cub}}%
}
\capttabbox{%
\footnotesize
  \begin{tabular}{llccc}
  \toprule
  &\multirow{2}{*}{Method}  & Acc. & Acc. & Acc.\\
   && Scenes & CUB & Flowers\\
  \toprule
    \textcolor{blue}{$\blacksquare$}& FT (CE)  &  $56.9\% \pm 1.1$ & $54.5\% \pm 2.6$& $89.3\% \pm 1.1$\\
    \textcolor{orange}{$\blacksquare$}& LwF &  $57.6\% \pm 1.5$ & $43.1\% \pm 2.9$& $85.3\% \pm 0.5$\\
    \textcolor{green}{$\blacksquare$}& $\text{EWC}_{\lambda:0.5k}$ &  $52.5\% \pm 1.1$ & $47.8\% \pm 2.5$& $85.9\% \pm 1.6$\\
    \textcolor{red}{$\blacksquare$}& $\text{EWC}_{\lambda:8k}$ &  $42.1\% \pm 1.5$ & $38.3\% \pm 0.9$& $79.1\% \pm 1.0$\\
    \textcolor{supconPurple}{$\blacksquare$}& FT (SupCon) &  $57.1\% \pm 1.2$ & $50.4\% \pm 1.0$& $85.3\% \pm 0.9$\\
    \toprule
  \end{tabular}
}{%
  \caption{\small Observed accuracy of the current task in the sequence ImageNet→Scenes→CUB→Flowers using ResNet architecture. Although $\text{EWC}_{\lambda:8k}$ attains relatively poor performance on the current task, it achieves the highest LP and observed accuracy for the previously seen tasks (see \cref{fig:imagenet-scenes-cub}). Moreover, the SupCon training achieves comparably high accuracy on the current task (even surpassing CE on Scenes) while suffering from relatively small representation forgetting (see \cref{fig:imagenet-scenes-cub}).
  \label{tab:imagenet-scenes-cub}\vspace{-25pt}}%
}
\end{floatrow}
\vspace{-25pt}
\end{figure*}
\vspace{-4pt}
\subsection{Exemplars and Fast Remembering}\vspace{-3pt}
Many CL methods utilize buffers of exemplars~\cite{rebuffi2017icarl} that are constantly updated. Typically, these samples are used repeatedly to train the model \cite{chaudhry2019continual}. In~\cite{rebuffi2017icarl, mai2021supervised}, the samples in the memory are also used to continuously estimate a class mean for old samples using the new representation. This is then used to construct a nearest mean of exemplars (NME) classifier, which can be seen as a fast way to construct a strong classifier that requires only a small amount of data. Instead of relying on the same exemplars during training and inference, one can use a small set of exemplars from the task distribution at the end of a task sequence to construct an NME based classifier in combination with a non-CL specific representation learning method such as SupCon~\cite{NEURIPS2020_d89a66c7}. Specifically the learner maintains a class mean for any class it has encountered. These class means are updated either by using a stored set of samples that is only used at inference or upon encountering an old task again, obtaining a small set of new samples to facilitate fast remembering. Notably, unlike the prior work on NME classifiers in continual learning, we don't suggest using exemplars as a rehearsal memory during the training, but as a method for fast remembering of class means for old tasks during evaluation time. This has the advantage of not increasing the computational complexity of training and not needing additional overhead in storing or retrieving samples, except at the end of a task sequence or upon re-encountering a task. Specifically, it can facilitate rapid remembering; in situations without prior data stored. Upon encountering a new task a model with minimal representation forgetting can rapidly adapt by updating just its class means.
\begin{figure}
    \centering
    \footnotesize LP vs. Observed Accuracy for SplitCIFAR100\par\medskip
    \vspace{-1em}
    \includegraphics[width=0.95\textwidth]{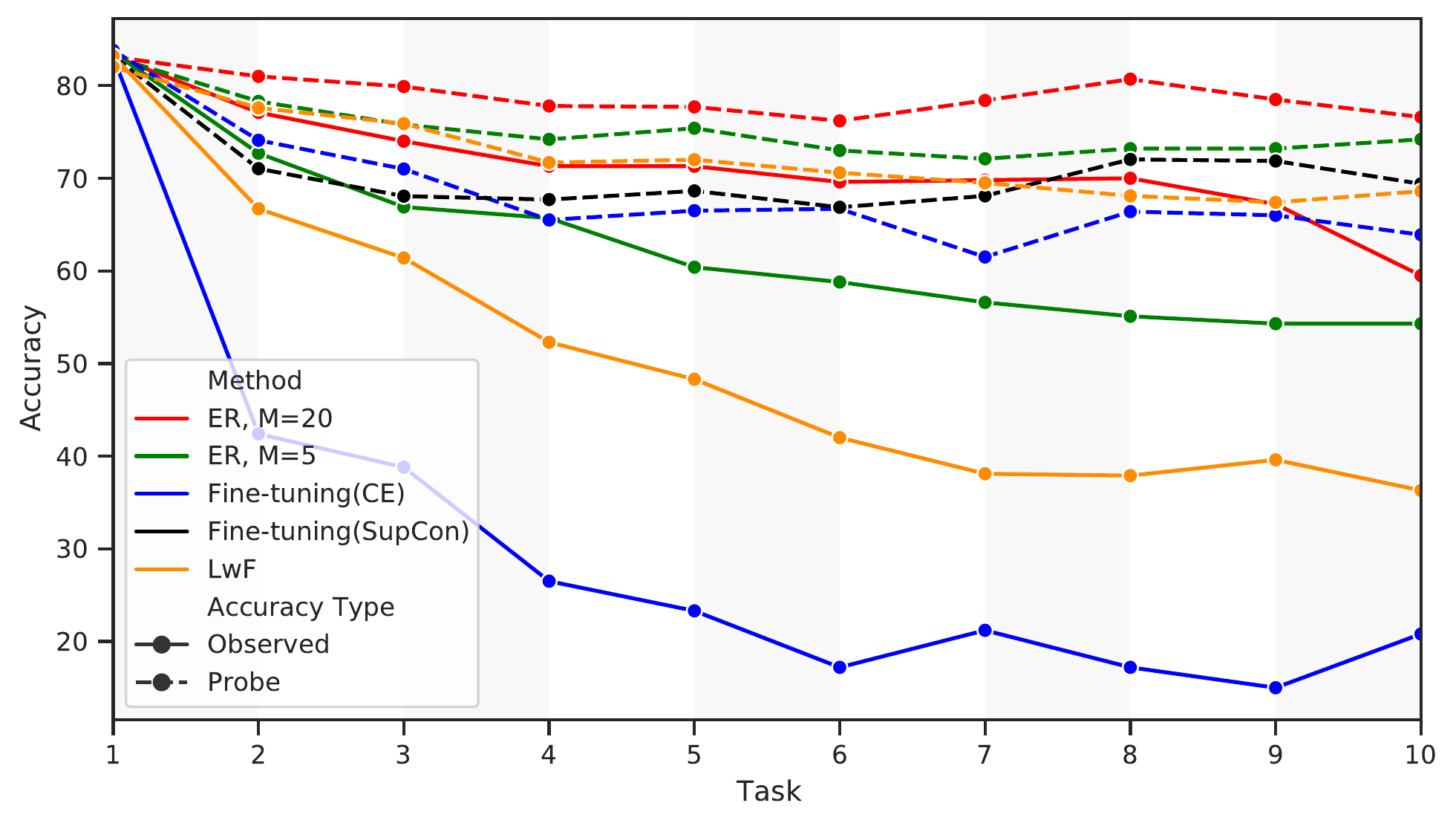}
    \vspace{-1em}
    \caption{10-Task SplitCIFAR100. We compare observed acc. and linear probe acc. Naively finetuning with CE does poorly if using the observed accuracy. However using the LP based evaluation we observe that performance gap to other methods is lower. Furthermore when finetuning is performed instead with the SupCon loss function LP performance can rival that of LwF.\vspace{-9pt}}
    \label{fig:cifar100_Task}
\end{figure}
\section{Experiments}
We perform evaluations in several CL scenarios, focusing on the task-incremental setting. The evaluations are based on LP and observed accuracy. Observed accuracy refers to the standard accuracy used in the CL literature. Specifically we measure observed accuracy, $A_{ij}$, as the accuracy of the model after step $i$ on the test data of task $j$. Similarly, the average observed accuracy at the end of the sequence is $\frac{1}{T}\sum_{t\in T}A_{T,t}$ as used in \eg~\cite{li2017learning}. Similarly we can measure the LP accuracy for each step $i$ and task $j$ as well as the average LP accuracy.   
\vspace{-9pt}
\paragraph{Datasets}
We use an ImageNet transfer setting based on~\cite{li2017learning}, a common SplitCIFAR100~\cite{krizhevsky2009learning} setting (split into 10 tasks), and reproduce the SplitCIFAR10 (split into 2 tasks) setting from~\cite{ramasesh2020anatomy}. Finally, to evaluate in very long task sequence regimes, we use a downsampled version of the entire ImageNet dataset (ImageNet32\cite{chrabaszcz2017downsampled}) split into 200 tasks of 5 classes each. For the ImageNet transfer setting, we use a sequence consisting of the standard  ImageNet (LSVRC 2012 subset)~\cite{ILSVRC15}, MIT Scenes~\cite{quattoni2009recognizing} for indoor scenes classification (5,360 samples over 67 classes), Caltech-UCSD Birds (CUB)~\cite{WelinderEtal2010} for classification of bird species (6,033 samples over 200 classes), and Oxford Flowers~\cite{Nilsback08} for flower classification (2,040 samples over 102 classes). The use of this sequence allows us to complement the standard long task sequence benchmarks with a more realistic and diverse larger scale sequence. Optimization hyperparameters for training  are detailed in the Appendix. 
\vspace{-8pt}\paragraph{Methods Compared} Our work focuses on evaluating naive finetuning based approaches using CE and SupCon~\cite{NEURIPS2020_d89a66c7} loss functions, as well as a set of representative CL methods. For regularization-based baselines, we consider two of the most popular methods, which do not utilize memory of any past samples: LwF~\cite{li2017learning} and EWC~\cite{huszar2017quadratic}. For rehearsal-based baselines, which continuously store past samples we focus on Experience Replay (ER). Indeed a number of recent works have illustrated that ER, particularly with increase in buffer size, is a strong baseline~\cite{chaudhry2019continual,ramasesh2020anatomy} and rivals or exceeds other rehearsal based methods such as iCaRL~\cite{rebuffi2017icarl} and GEM~\cite{lopez2017gradient}. Hence we use ER with both a small buffer, $M=5$ samples per class, and a relatively large buffer, $M=20$ samples per class.

\begin{figure}
\centering
\footnotesize Observed and LP Accuracy SplitMiniImageNet\par\medskip
\vspace{-1em}
\includegraphics[width=0.95\textwidth]{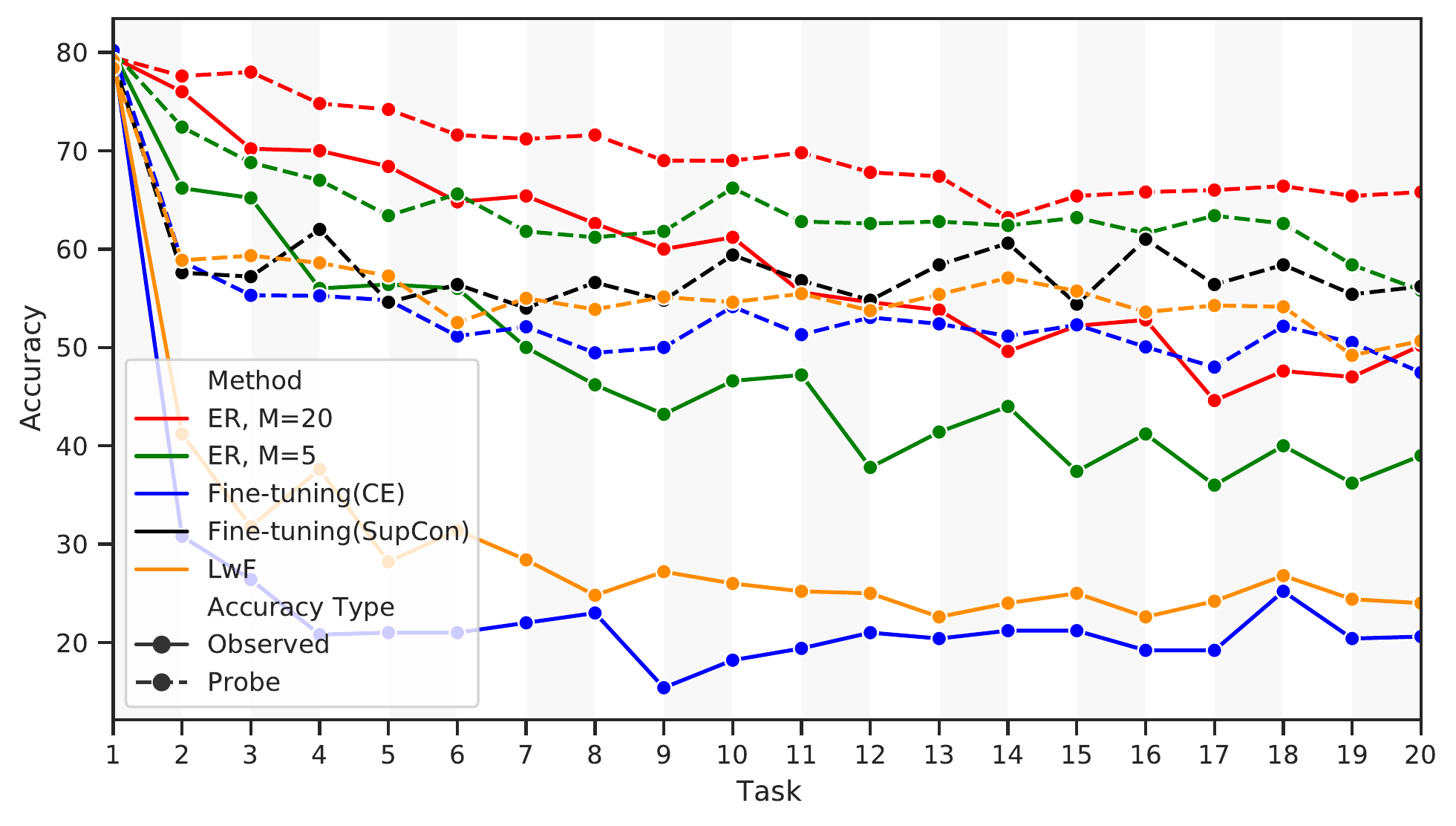}
\vspace{-1em} 
    \caption{20-Task SplitMiniImageNet. We compare observed accuracy and Linear probe accuracy for Task 1 data. Naively finetuning with CE as well as the LwF method does poorly if using the observed accuracy. However using the LP based evaluation we see that performance gap to other methods is less significant. LwF performs similar to finetuning with SupCon. For this Longer task sequence ER with large buffer,  performance decays towards the end of the sequence, while SupCon stays flat.\vspace{-12pt}}
    \label{fig:miniimagenet_Task}
\end{figure}

\makeatletter \g@addto@macro\@floatboxreset\centering \makeatother 
\begin{figure}
    \centering
    \footnotesize LP Accuracy for SplitCIFAR100\par\medskip
\vspace{-1em}
\includegraphics[width=0.95\textwidth]{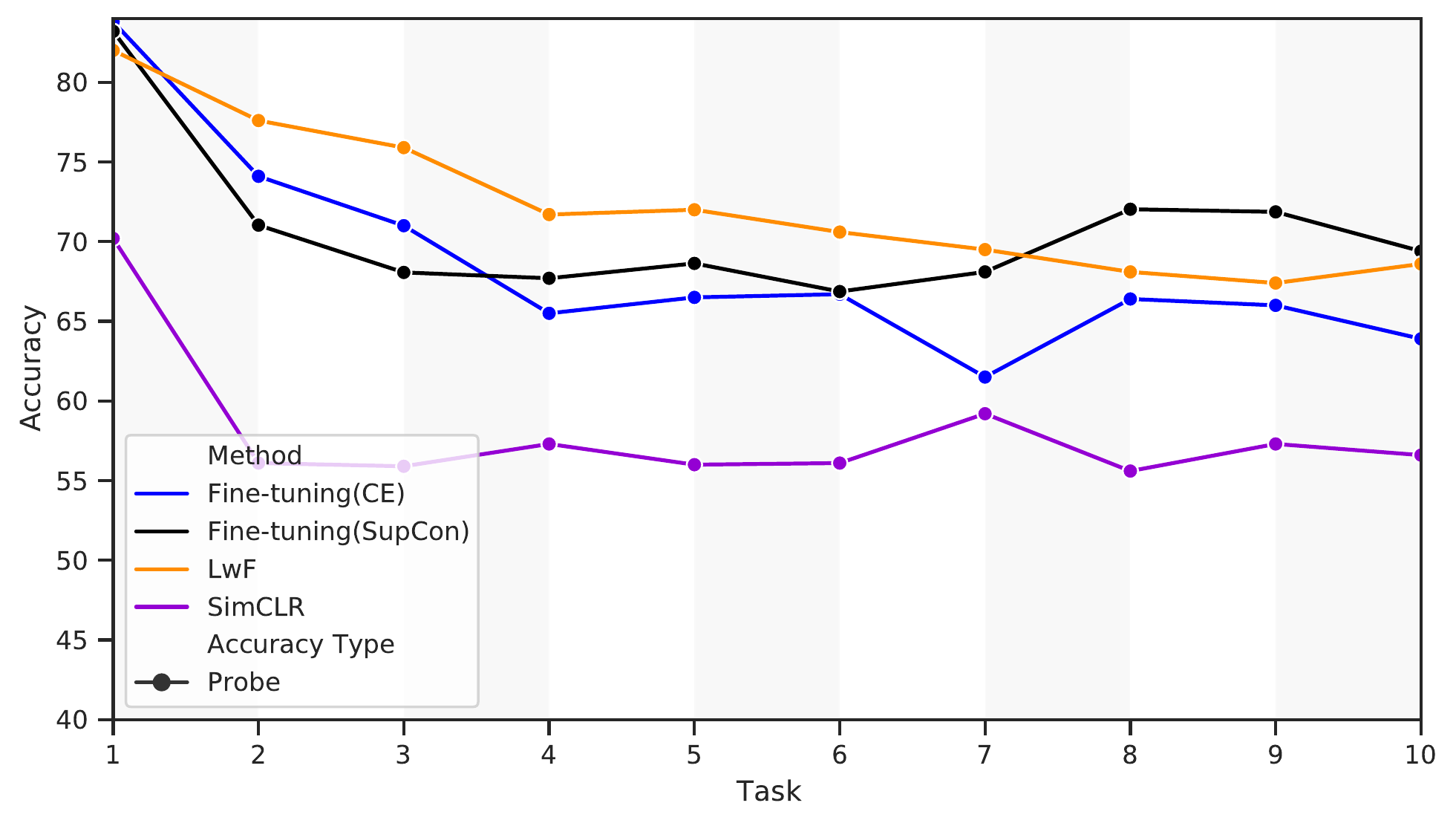}
\vspace{-1em} 
    \caption{SplitCIFAR100 comparison of unsupervised Linear Probe accuracies on task 1 with supervised finetuning CE and SupCon as well as LwF. We observe that LwF and CE based finetuning can decay over time, while the unsupervised learning (SimCLR) has an initial drop and stays relatively flat.\vspace{-12pt}}
    \label{fig:miniimagenet_unsup}
\end{figure}

\subsection{Observed vs LP accuracy} 
\label{sec:observed-vs-lp-acc}
In this section we study the observed vs LP accuracy for various task sequences and methods, in both supervised and unsupervised settings.\vspace{-12pt}
\paragraph{ImageNet Transfer}We consider models trained on the large ImageNet data and subsequently applied to different tasks in the sequence. We take the setting of~\cite{li2017learning}, which considers the ImageNet~\cite{ILSVRC15} transfer to various datasets, in particular CUB~\cite{wah2011caltech}, and Scenes~\cite{quattoni2009recognizing}. We extend this setting by including Flowers~\cite{Nilsback08} in the task sequence.
To reduce the computation of experiments we do random resize crops to 64$\times$64 and utilize ResNet-18 for these experiments. Additional results further confirming our observations with larger crop size are given in the Appendix. As mentioned earlier, in addition to LwF, we also examine EWC~\cite{huszar2017quadratic} 
under two conditions: (a) large $\lambda$ value ($8k$), so the network is inclined to preserve the knowledge important to the previous tasks, and (b) small $\lambda$ value ($0.5k$), so the network is encouraged to perform competitively on the current task. 

\begin{figure*}[t!]
\vspace{-1em}
\centering
\footnotesize LP Accuracy ImageNet 200 Task Sequence\par\medskip
\vspace{-1em} 
\includegraphics[width=0.75\textwidth]{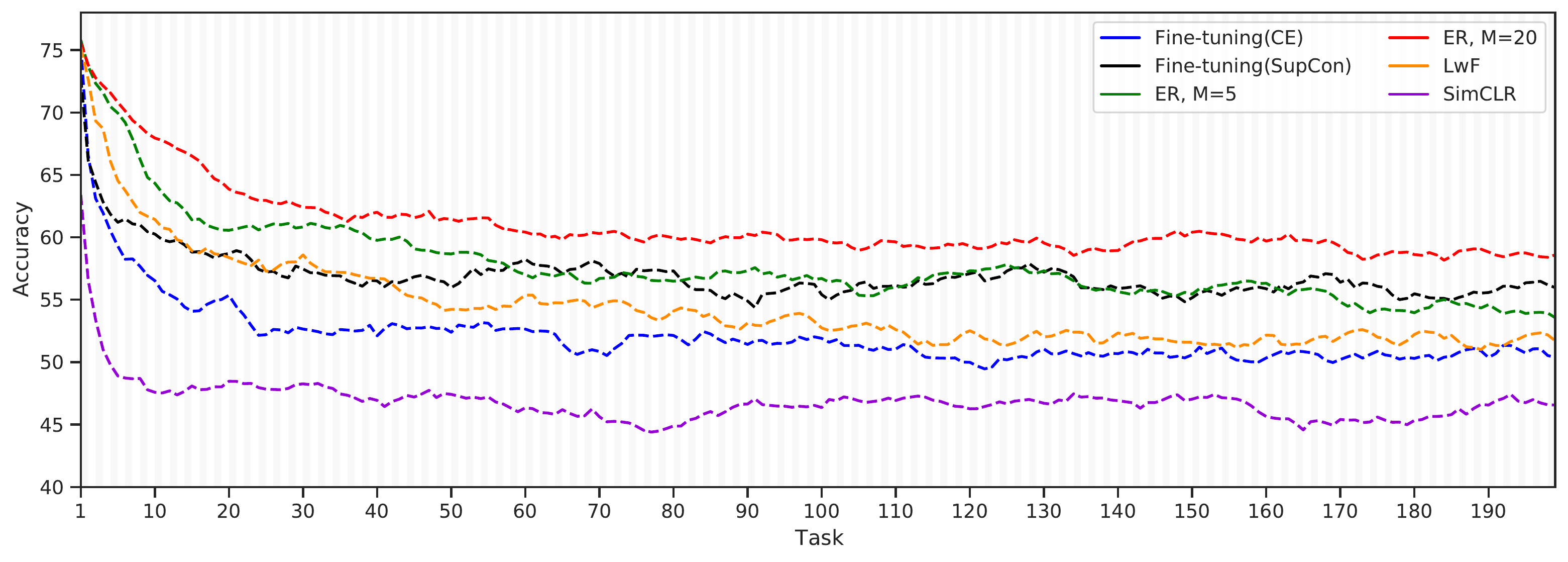}
\vspace{-1em}
    \caption{200-Task ImageNet32. We compare Linear probe accuracy for Tasks 1 data over the whole sequence. As the model observes the later tasks of the sequence, the performance of finetuning with CE reaches LwF, and finetuning with SupCon outperforms ER with 5 samples per class.\vspace{-12pt}}
    \label{fig:imagenet32}
\end{figure*}

We report observed accuracy and LP accuracy on ImageNet validation set as the model is trained on the task sequence ImageNet $\to$ Scenes $\to$ CUB $\to$ Flowers (see~\cref{fig:imagenet-scenes-cub}). We also report the observed accuracy on the current task in~\cref{tab:imagenet-scenes-cub}.
Our evaluation reveals that although the forgetting in terms of the traditional measure is high for finetuning compared to LwF (as shown in~\cite{li2017learning}), the LP accuracy of these methods suggest less drastic forgetting. Furthermore, the LP performance across finetuning and other methods is not as drastically different as their respective observed accuracies are. We see that the LP accuracy of SupCon based finetuning, which has no explicit control for forgetting, outperforms LwF, a method specifically designed for CL. It also closely tracks the performance of $\text{EWC}_{\lambda:0.5k}$, while outperforming on the current task performance. Indeed, as we can see in~\cref{tab:imagenet-scenes-cub}, SupCon training achieves comparably high accuracy on the current task (even surpassing CE finetuning on Scenes) with relatively small representation forgetting (see \cref{fig:imagenet-scenes-cub}).\vspace{-10pt}
\paragraph{SplitCIFAR100 and SplitMiniImageNet} We now consider the SplitCIFAR100 with 10 tasks of 10 classes each and the 20 task SplitMiniImageNet setting. We show in \cref{fig:cifar100_Task} and \cref{fig:miniimagenet_Task} the performance on the first task throughout the sequence for both settings. %
Similar to the previous case we see: for finetuning with CE the LP based evaluation shows much milder forgetting than observed accuracy. When finetuning with SupCon, LP performance drops initially but then stays relatively flat and even increases, suggesting that positive backward transfer is occurring. Overtime, SupCon outperforms the LwF~\cite{li2017learning} approach without any specific CL based control in both task sequences. It also obtains performance that becomes close to ER with 5 samples per class without access to any previous data. For the longer MiniImageNet task sequence we see that over time even the strong ER based baselines which train on old data demonstrate a reduced LP performance, while finetuning baselines remain relatively flat and even increase in the case of SupCon. This suggests that in very long sequences, finetuning baselines can be competitive to the more computationally expensive CL methods.

Utilizing SplitCIFAR100 we also consider the unsupervised representation learning case where linear probes follow naturally as a common evaluation setting. The literature on evaluation of continual learning methods in the unsupervised setting is limited. Hence we directly compare unsupervised and supervised approaches in their representation learning ability when presented with the same task sequences.  We focus on the SimCLR loss and evaluate LP performance in comparison to other methods in \cref{fig:miniimagenet_unsup}. We can see in \cref{fig:miniimagenet_unsup} that the initial LP performance is lower for SimCLR compared to the supervised losses. This is natural as it does not have access to the labeled data. Despite the higher starting accuracy, LwF and finetuning with CE show a decay that continues in the first several tasks following task 1. On the other hand SimCLR decays at the first step but then remains nearly flat over the rest of the sequence, showing a strong resistance to representation forgetting after this inital drop. However SupCon, which utilizes a loss similar to SimCLR in the supervised setting, shows the best of both worlds, has an initial drop and then illustrates gradual backward transfer properties.

\newcommand\MyLBrace[2]{\Large
 \rule{0pt}{#1}\right\text{#2} \Big\{}
\begin{table*}[t!]
\vspace{-1em}
  \caption{Final Accuracy of 10 task SplitCIFAR100 sequence with variable width and depth in the offline setting. $M$ indicates the number of samples per task used in the ER buffer. We observe that simple  finetuning and LwF baselines show large forgetting, which do not improve significantly with width or depth. On the other hand, the LP evaluation reveals that representation quality for finetuning and LwF becomes closer to strong CL methods that store samples and also use more compute such as ER. Furthermore, LP evaluations reveal LwF (which does not store prior data) with wider models can rival ER}
  \label{tab:CIFARLongTask_Offline}
  \centering
\footnotesize
  \begin{tabular}{c|c|ccccc}

	\multicolumn{2}{c}{ }		&	Task 1	&	Obs. Acc. Task 1	&	Task 1 LP	&	LP Acc. All 	&	Avg. Obs. 	\\
\multicolumn{1}{c}{}		&	\multicolumn{1}{c}{}	&	Acc.	&	at T=10	&	T=10	&	T=10	&	Acc.	\\\hline
	&	RN18, Width=32	&	82.2	&	20.8	&	64.8	&	70.8	&	35.5	\\	
Finetuning(CE)	&	RN18, Width=128	&	83.3	&	21.2	&	70.5	&	74.2	&	36.8	\\	
	&	RN101, Width=32	&	82.9	&	19.8	&	67.9	&	72.4	&	35.9	\\	\hline
	&	RN18, Width=32	&	82.7	&	52.1	&	74.2	&	75.7	&	54.8	\\	
ER-M5	&	RN18, Width=128	&	83.6	&	54.8	&	75.6	&	77.3	&	55.4	\\	
	&	RN101, Width=32	&	83	&	50.9	&	74.5	&	76.1	&	51.6	\\	\hline
	&	RN18, Width=32	&	82.4	&	61.3	&	76	&	76.4	&	65.2	\\	
ER-M20	&	RN18, Width=128	&	83.2	&	63.5	&	78.8	&	80.1	&	67	\\	
	&	RN101, Width=32	&	82.9	&	60.7	&	77.1	&	77.5	&	63.9	\\	\hline
	&	RN18, Width=32	&	82.1	&	36.2	&	70.1	&	73.4	&	47.7	\\	
LwF	&	RN18, Width=128	&	83.9	&	37.7	&	74.8	&	76.7	&	49.1	\\	
	&	RN101, Width=32	&	82.5	&	35.5	&	71.0	&	74.6	&	46.3	\\	

  \end{tabular}
\end{table*}

\begin{table*}[t!]
  \caption{Final Accuracy of 10 task SplitCIFAR100 sequence in the Online Setting. LP evaluations show that width  substantially improves online representation learning, while observed Avg Accuracies suggest it decreases. Increasing depth on the other hand appears to be less effective in the online setting.\vspace{-12pt}}
  \label{tab:CIFARLongTask_Online}
  \centering
\footnotesize
  \begin{tabular}{c|c|ccccc}

	\multicolumn{2}{c}{ }&	Task 1	&	Obs. Acc. Task 1	&	Task 1 LP	&	LP Acc. All 	&	Avg. Obs. 	\\
		\multicolumn{2}{c}{ }		&	Acc.	&	at T=10	&	T=10	&	T=10	&	Acc.	\\\hline
	&	RN18, Width=32	&   18.6	&	12.2	&	39.8	&	36.4	&	22.3	\\
Finetuning(CE)	&	RN18, Width=128	&   19.4  	&	12.7	&	42.3	&	41.7	&	19.8	\\
	&	RN101, Width=32	&	14.6	&	11.8	&	28.2	&	29.4	&	14.5	\\\hline
	&	RN18, Width=32	&	18.8	&	27.3	&	36.0	&	40.1	&	33.8	\\
ER-M5	&	RN18, Width=128	&	19.5	&	28.9	&	54.7	&	47.9	&	31.6	\\
	&	RN101, Width=32	&   15.0	&	24.7	&	37.1	&	30.4	&	24.3	\\\hline
	&	RN18, Width=32	&	18.4	&	32.0	&	46.8	&	43.5	&	34.7	\\
ER-M20	&	RN18, Width=128	&   20.0	&	31.8	&	51.2	&	50.7	&	32.5	\\\
	&	RN101, Width=32	&	14.5	&	25.4	&	36.5	&	33.9	&	24.3	\\\hline
	&	RN18, Width=32	&	18.5	&	13.4	&	29.5	&	36.0	&	22.7	\\
LwF	&	RN18, Width=128	&	19.7	&	18.3	&	34.6	&	39.1	&	22.1	\\
	&	RN101, Width=32	&	14.8	&	11.1	&	25.4	&	22.8	&	16.8	\\\hline

  \end{tabular}
\end{table*}

\vspace{-5pt}
\paragraph{200 Task Sequence - SplitImageNet32}
We now consider a much longer sequence than typically studied in the literature to allow us to observe whether the trends we have seen so far continue to hold. Using Imagenet32 we construct  200 tasks of 5 classes each. \cref{fig:imagenet32} shows the performance on the first task throughout the whole sequence. We see that in a very long sequence of tasks, the previous trends are kept. Specifically, we see that as the model reaches the later stages of the sequence, finetuning with CE reaches LwF, and finetuning with SupCon outperforms the competitive baseline of ER with a small buffer \textit{without} access to buffer data. Furthermore, we observe as in the previous section that SimCLR performance stays flat. In the supplementary material we also demonstrate that this pattern is not limited to the first task but is maintained for other tasks along the sequence.

\subsection{Effects of Increased Model Capacity}\vspace{-3pt}
\label{sec:effect-of-increased-model-capacity}
Next, we use linear probes to evaluate the effect of increased model width and depth. Recently~\cite{anonymous2022effect} has suggested that increased model size must be strictly combined with pre-training in order to get increased robustness to catastrophic forgetting. We revisit this in the context of both wider and deeper models on a SplitCIFAR100 sequence of 10 tasks with 10 classes each. \Cref{tab:CIFARLongTask_Offline} shows the results using a Resnet18 with a much wider model (128 vs. 32) and then a much deeper model (101 layers). We report both the LP accuracy of task 1 at the end of the sequence and the average of LP accuracies for all the tasks at the end of the sequence. 

First, we can see that as in the other cases, the LP accuracy of finetuning is higher than observed accuracy, suggesting that forgetting is less catastrophic than what is indicated by observed accuracy. Secondly, we see that finetuning evaluated using the observed accuracy is particularly deceptive in revealing how the model representations changes with increasing capacity. The observed  and accuracy of task 1 are relatively close despite increasing capacity (wider or deeper) while the corresponding LP accuracies show substantial gaps. Using observed accuracy one would conclude that increasing width and capacity of the model without applying any CL specific method does not reduce forgetting. This is consistent with the observations of~\cite{anonymous2022effect}, which evaluates only on observed accuracy. However, if we observe the LP accuracy, it reveals a more clear picture of what occurs at the representation level, suggesting that larger models can indeed reduce forgetting even when trained from scratch without explicit control of forgetting. Moreover, we see that at the representation level as model capacity increases, naive finetuning becomes much closer in performance to costly (and under privacy constraints unusable) CL methods such as ER, which require more compute and memory. 

In comparing depth and width we also see some key distinctions - increasing width appears to help more than increasing depth. For ER we also see that increasing depth yields a lower observed accuracy, while the LP evaluation suggests the representations are similar.
Similarly, in~\cref{tab:CIFARLongTask_Online} we report the results for the online task-incremental setting~\cite{lopez2017gradient,chaudhry2019continual}. In this setting, momentum tends to be detrimental to performance, thus we use a fixed learning rate of $0.01$ with no momentum. We see  similar behavior to the previous case, the larger models can end up appearing to do worse if we consider observed accuracy, but perform better using LP evaluation. Wider models appear to do particularly well in the online setting while deeper models have degraded LP accuracy in this setting. Finally we see that LwF which is a regularization method performs poorly in this setting. Indeed regularization based methods do poorly in the online setting. This suggests that amongst methods without access to a replay buffer, finetuning may provide the best representation learning.

\subsection{Low-Cost Remembering with SupCon}\vspace{-3pt}
The observed low representation forgetting properties of finetuning with SupCon loss suggest if we can approximate a classifier using it's representation it would allow for low cost remembering upon encountering a previously observed task. We thus evaluate the use of the NME in combination with SupCon.  As discussed in Sec 3.4 such an approach allows a simpler alternative to ER methods and moreover facilitates fast remembering not relying on a buffer and repeatedly training the model with old samples. We use the SplitCIFAR100 dataset to compare against several  CL specific methods such as LwF and ER in \cref{tab:SplitCIFAR100-NME}. We use a memory with $M=5$ samples per class for this. We chose the exemplars at random to simulate re-encountering an old task. We observe that just applying the simple approximation with a small number of samples allows for a rapid recovery of the performance with the finetuning approach alone, exceeding the performance of LwF on overall accuracy and task 1 accuracy. The overall performance is close to that of ER with the same  memory size and slightly below the ER performance on task 1 at the end of the sequence. On the other hand ER requires samples to be available during the entire training sequence, requires the addition of extra algorithmic elements specifically to control forgetting, and uses substantially more compute ($\approx 2 \times$ that of the finetuning step in this case). 

\setlength{\tabcolsep}{2pt}
\begin{table}[tb]
    \caption{Final Accuracy of 10-task SplitCIFAR100 sequence comparing only the observed accuracy and SupCon+NME. Supcon+NME gives superior performance to CL specific methods such as LwF and nearly matches the performance of ER with a similar memory size while not needing access to the memory during task training.\vspace{-15pt}}
  \label{tab:SplitCIFAR100-NME}
  \centering
  \begin{tabular}{lcc}
    \toprule
    \multirow{2}{*}{Method}   &   Obs Acc. Task 1   &   Avg. Obs.\\
    &   at T=10 & Acc.\\
    \midrule
    Finetuning(CE) &  $20.8\%$ &  $35.5\%$\\
    ER-M5 &  $52.1\%$ &  $54.8\%$\\
    ER-M20 &  $61.3\%$ &  $65.2\%$\\
    LwF &  $36.2\%$ &  $47.7\%$\\
    Finetune(SupCon) + NME-M5 &  $48.0\%$ &  $53.9\%$\\
   \bottomrule
  \end{tabular}
  
\end{table}
\subsection{Depth-wise Probes and Comparison to CKA}
\label{sec:depthwise-probe-and-comparison-to-cka}

We consider a 2-task SplitCIFAR10 setting from~~\cite{ramasesh2020anatomy}. We use the same models and training procedures and subsequently evaluate forgetting. In~\cref{tab:SplitCIFAR10-studies}, we study the shift in representations of each block of the network by measuring the performance of LP on task 1 before and after training the network on task 2.

\begin{table}[tb]
  \caption{Representation forgetting of Task 1 measured via optimal linear probes (LP) on ResNet and VGG. The Accuracy degradation of LP trained on activations of stages (blocks of convolutions) before and after observing Task 2 suggests that the representations are still highly useful for Task 1 despite training on Task 2. *Note CKA results are taken from~\cite{ramasesh2020anatomy} for comparison.\vspace{-15pt}}%
  \label{tab:SplitCIFAR10-studies}
  \centering
  \small
  \begin{tabular}{lccrc}
    \toprule
    \multicolumn{5}{c}{ResNet: Network Acc. on T-1 after T-2 training: 63.64\%} \\
    \cmidrule(r){1-5}
    \multirow{2}{*}{Block}   &   LP Acc. & LP Acc.  & \multirow{2}{*}{$\Delta$ Acc.} & \multirow{2}{*}{CKA*}\\
    &   \hspace*{1em} Post T-1 \hspace*{1em} & \hspace*{1em} Post T-2 \hspace*{1em}\\
    \midrule
    B-0 &  $63.54\%$ &  $64.62\%$ & $+1.08\%$ & $0.97$\\
    B-1 &  $68.24\%$ &  $69.50\%$ & $+1.26\%$ & $0.93$\\
    B-2 &  $71.62\%$ &  $71.34\%$ & $-0.28\%$ & $0.88$\\
    B-3 &  $77.64\%$ &  $76.52\%$ & $-1.12\%$ & $0.78$\\
    B-4 &  $80.06\%$ &  $78.98\%$ & $-1.08\%$ & $0.31$\\
    B-5 &  $85.82\%$ &  $80.10\%$ & $-5.72\%$ & $0.22$\\
    \toprule
    \multicolumn{5}{c}{VGG: Network Acc. on T-1 after T-2 training: 57.88\%} \\
    \cmidrule(r){1-5}
    B-0 &  $67.94\%$ &  $66.86\%$ & $-1.08\%$ & $0.95$\\
    B-1 &  $73.60\%$ &  $72.52\%$ & $-1.08\%$ & $0.93$\\
    B-2 &  $78.58\%$ &  $75.68\%$ & $-2.90\%$ & $0.85$\\
    B-3 &  $81.54\%$ &  $75.48\%$ & $-6.06\%$ & $0.66$\\
   \bottomrule
  \end{tabular}
  \vspace{0pt}
\end{table}

First we see that the observed accuracy decreases from $85\%$ to $63\%$, suggesting large degradation in performance and large forgetting. However, following the optimal classifier evaluation protocol the accuracy degradation is seen to be only $5.7\%$, without any CL method applied to control forgetting.  This suggests that the representations are still highly useful for Task 1 despite training on Task 2. Second, similar to~\cite{ramasesh2020anatomy} we note that the forgetting is concentrated at the top layers. Indeed early layers in the network experience almost no representation forgetting and in some cases improve their usefulness with regards to Task 1. 
Ramasesh \etal's~\cite{ramasesh2020anatomy} analysis also showed forgetting occurring in early layers to a lower degree than in higher layers and suggested that forgetting is extreme in the upper layer representations. Specifically, the authors measured linear CKA~\cite{kornblith2019similarity} performance between layers (given in \cref{tab:SplitCIFAR10-studies}) showing that this similarity metric dropped progressively from close to 1 to 0.2 for both ResNet and VGG models. However, our evaluation suggests that forgetting does not exist in lower layers and also the loss in information is less catastrophic at higher layers than suggested by~\cite{ramasesh2020anatomy}. 

\vspace{-4pt}\section{Conclusion}
We have highlighted the importance of evaluating representations and not just task accuracy in CL. Our results suggest a) representation forgetting under naive finetuning in supervised settings is not as catastrophic as other metrics suggest b) We demonstrate that without evaluation of features the effects of model size on forgetting and representation learning will be misinterpreted. c) We show that the self-supervised SimCLR loss and supervised SupCon loss have lesser representation forgetting in long tasks sequences, maintaining or increasing performance on early tasks. These results open up potential new directions for approaches in continual learning. One such direction of using memories that are not available to the learner during training is evaluated with promising initial results.

\vspace{-0.2em}
\paragraph{Limitations and Future Work}
Our work focuses on comparing linear probe performance as a proxy of knowledge retained from past tasks. However, task performance may not be the only criteria to fully evaluate knowledge retention related to past data. Another limitation of our work is that it currently focuses on the task-incremental setting and does not consider the important class-incremental setting, a subject for future studies. Finally though our work studies a diverse task sequence ImageNet $\to$ Scenes $\to$ CUB $\to$ Flowers, to fully understand the behavior of representation forgetting, results over more distant tasks may be needed (e.g. ImageNet $\to$ Sketch Images~\cite{sangkloy2016sketchy,ha2017neural})  %

{\small
\bibliographystyle{ieee_fullname}
\bibliography{egbib}
}
\newpage
\onecolumn
\section*{APPENDIX}
\subsection{Training Details}

\paragraph{ImageNet $\to$ Scenes $\to$ CUB $\to$ Flowers} we use images downsampled and cropped at 64$\times$64 (224$\times$224 size results are shown in \cref{sec:reproducing-li}) in this task sequence. We train a ResNet-18~\cite{he2016deep}, using AdamW~\cite{loshchilov2017decoupled} optimizer with a learning rate (LR) of $1e\texttt{-}5$ and a weight decay of $5e\texttt{-}4$. We first train the model on ImageNet via (1) CE loss, or (2) SupCon loss for 500 epochs. Next, for each task in the sequence, we train the model via early stopping~\cite{caruana2001overfitting}, and report the observed accuracy over the test set (see \cref{tab:imagenet-scenes-cub}).

The LP classifiers are trained using the same optimizer and weight decay value. However, for faster convergence, in this setting, we use a cosine LR scheduler~\cite{loshchilov2016sgdr} with an initial LR of $1e\texttt{-}4$. The results of LP classifiers are reported on the ImageNet validation set. All the training images in this section undergo, random crop, random horizontal flip, and color jitter of 0.5.

\vspace{-11pt}
\paragraph{SplitCIFAR100, MiniImageNet, ImageNet32}
For our SplitCIFAR100 10-task sequence, MiniImageNet 20-task sequence, and ImageNet32 200-task sequence we use SGD optimizer with LR of 0.05, momentum of 0.9, and weight decay of $1e\texttt{-}4$. We train the models for 50 epochs for SplitCIFAR100 and MiniImageNet, and for 80 epochs for ImageNet32. The augmentation pipeline consists of random crop, random horizontal flip, and color jitter of 0.5. For the LP classifiers, we train each for 20 epochs, using AdamW~\cite{loshchilov2017decoupled} optimizer with a LR of $1e\texttt{-}3$ and a weight decay of $5e\texttt{-}4$.

For the SplitCIFAR10 2-task sequence from~\cite{ramasesh2020anatomy} we use the hyper-parameters and training conditions mentioned in~\cite{ramasesh2020anatomy} for both VGG~\cite{simonyan2014very} and ResNet~\cite{he2016deep} architectures. For the LP classifiers, we train each for 70 epochs, using AdamW~\cite{loshchilov2017decoupled} optimizer with a LR of $1e\texttt{-}3$ and a weight decay of $5e\texttt{-}4$. For LP training we use data augmentation in all cases, except the SplitCIFAR10 to obtain the most accurate measure of the model's ability to linearly separate the data of interest. For SplitCIFAR10 2-task we choose not to use any data augmentation because the source result (\cite{ramasesh2020anatomy}) does not rely on data augmentation for training. We opt not to use data augmentation in our LP evaluation for a fair comparison.

\subsection{ImageNet$\to$ Scenes$\to$ CUB with 224 $\times$ 224: Reproducing Li et al.}
\label{sec:reproducing-li}
\begin{figure*}[!ht]
\TopFloatBoxes
\begin{floatrow}
\ffigbox[0.65\textwidth]{%
    \includegraphics[width=\linewidth]{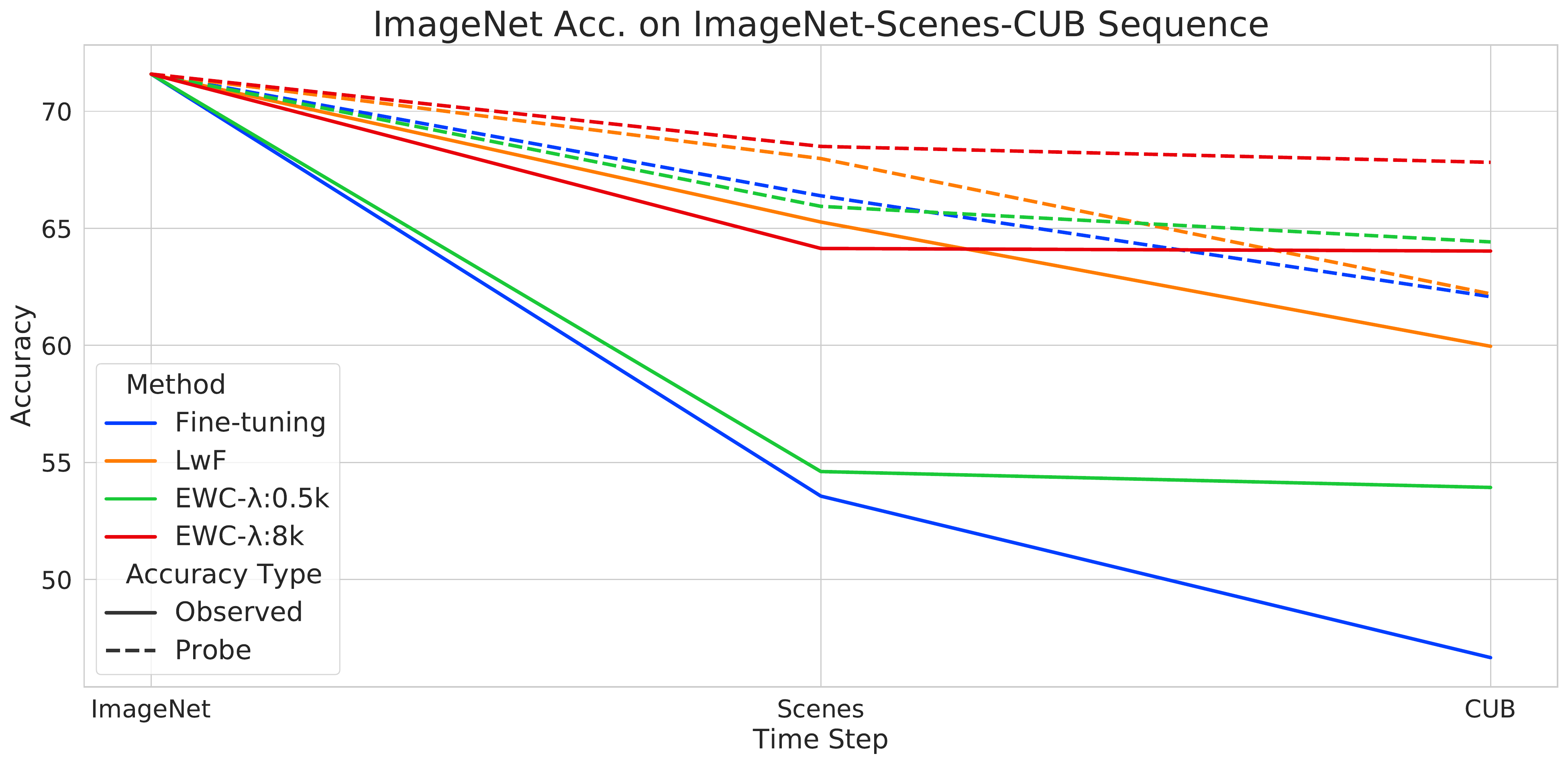}
}{%
  \caption{\small Performance on ImageNet during the transfer sequence (ImageNet→Scenes→CUB) using VGG-16. We observe that although observed accuracy heavily degrades, the LP accuracy for finetuning does not decay as drastically and can rival LP accuracy of methods such as LwF and EWC. We evaluate methods which do not rely on storing data from Task 1 to replay during training. Note EWC with $\lambda=8k$ is the best performing method in terms of LP and observed accuracy, however it does not perform well on the current task (see \cref{tab:larg-imagenet-scenes-cub}).
  \label{fig:large-imagenet-scenes-cub}}%
}
\capttabbox{%
\small
  \begin{tabular}{llcc}
  \toprule
  &\multirow{2}{*}{Method}  & Acc. & Acc. \\
   && Scenes & CUB\\
  \toprule
    \textcolor{blue}{$\blacksquare$}& Finetuning  &  $74.70\%$ & $74.39\%$\\
    \textcolor{orange}{$\blacksquare$}& LwF &  $74.78\%$ & $75.23\%$\\
    \textcolor{green}{$\blacksquare$}& $\text{EWC}_{\lambda:0.5k}$ &  $74.70\%$ & $74.72\%$\\
    \textcolor{red}{$\blacksquare$}& $\text{EWC}_{\lambda:8k}$ &  $72.69\%$ & $71.44\%$\\
    \toprule
  \end{tabular}
}{%
  \caption{\small Observed accuracy of the current task in the sequence ImageNet→Scenes→CUB using VGG-16 architecture. Although $\text{EWC}_{\lambda:8k}$ attains relatively poor performance on the current task, it achieves the highest LP and observed accuracy for the previously seen tasks (see \cref{fig:large-imagenet-scenes-cub}).
  \label{tab:larg-imagenet-scenes-cub}}%
}
\end{floatrow}
\end{figure*}

For ease of experiments, in \cref{sec:observed-vs-lp-acc} we have used a lower resolution, 64$\times$64, a light ResNet-18 model, and a modern rapid training scheduler. On the other hand in this section we reproduce the results of the original paper (Li \etal)~\cite{li2017learning} using a VGG-16 model and 224$\times$224 input size and then carry out our LP analysis further confirming our observations. 
We take the setting of~\cite{li2017learning}, which considers the ImageNet~\cite{ILSVRC15} transfer to various datasets, in particular CUB~\cite{wah2011caltech}, and Scenes~\cite{quattoni2009recognizing}. We use the same model architecture (VGG-16) and training procedures described in ~\cite{li2017learning}, which proposes LwF, closely reproducing their ImageNet $\to$ Scenes as the first step in the sequence (see \cref{tab:lwf-vs-ours}). We also include an EWC baseline under two conditions: (a) large $\lambda$ value ($\lambda=8k$), so the network is inclined to preserve the knowledge important to the previous tasks, and (b) small $\lambda$ value ($\lambda=0.5k$), so the network is encouraged to perform competitively on the current task. The results are shown in Figure~\ref{fig:large-imagenet-scenes-cub} and Table~\ref{tab:larg-imagenet-scenes-cub}.

We first note that our results for the first task switch are consistent with those reported in~\cite{li2017learning} (see~\cref{tab:ImageNet-based-studies}). \cref{fig:large-imagenet-scenes-cub} reveals that although the forgetting in terms of the traditional measure is high for finetuning compared to LwF (as shown in~\cite{li2017learning}), the LP accuracy of these methods suggest a much less drastic forgetting. Furthermore, the LP performance across finetuning and other methods is not as drastically different as their respective observed accuracies are.  Indeed, we observe that on the third task, finetuning outperforms LwF in representation forgetting on ImageNet.
Similarly EWC does not clearly outperform naive finetuning. For example, if using one hyper-parameter for the regularization term the performance closely tracks finetuning. On the other hand using $\lambda=8k$ we observe the best LP performance on ImageNet through the task sequence but degraded current task performance as seen in~\cref{tab:larg-imagenet-scenes-cub}.

\begin{table}[!ht]
\vspace{-5pt}
\RawFloats
\parbox{.48\linewidth}{
  \centering
  \small
  \begin{tabular}{lrc}
    \toprule
    &\multicolumn{2}{c}{ImageNet (T-1) $\to$ CUB (T-2)}\\
    \cmidrule(r){2-3}
    & Obs. T-2 Acc. & Obs. T-1 Acc.\\
    \midrule
    Finetune-\cite{li2017learning}   &  $73.1\%$   &   $50.7\%$\\
    Finetune-Ours         &  $74.5\%$   &   $50.9\%$\\
    LwF-\cite{li2017learning}   &  $72.5\%$   &   $60.6\%$\\
    LwF-Ours         &  $75.7\%$   &   $63.6\%$\\
    \midrule
    &\multicolumn{2}{c}{ImageNet (T-1) $\to$ Scenes (T-2)}\\
    \midrule
    Finetune-\cite{li2017learning}    &   $74.6\%$    & $62.7\%$  \\
    Finetune-Ours         &   $74.7\%$    &   $53.6\%$\\
    LwF-\cite{li2017learning}   &  $74.9\%$   &   $66.8\%$\\
    LwF-Ours         &  $74.8\%$   &   $65.3\%$\\
  \bottomrule
  \end{tabular}
  \caption{Reproduction of the results reported in~\cite{li2017learning}. Note that we observe a slight difference in our reproduced results due to stochasticity of training neural networks, and removing the warm-up step.}
  \label{tab:lwf-vs-ours}
}
\hfill
\parbox{.48\linewidth}{
  \centering
  \begin{tabular}{lclr}
    \toprule
    \multicolumn{4}{c}{Observed Acc. on ImageNet: 71.59\%}\\
    \toprule
    &\multicolumn{3}{c}{ImageNet (T-1) $\to$ CUB (T-2)}\\
    \cmidrule(r){2-4}
    & T-1 Acc. @ T-2 & LP Acc.  @ T-2 & T-2 Acc.\\
    \midrule
    FT  &  $50.89\%$ & $64.49\%$ & $74.51\%$\\
    LwF &  $63.58\%$ & $67.23\%$ & $75.65\%$\\
    $\text{EWC}_{\lambda:8k}$ &  $60.28\%$ & $67.46\%$ & $72.70\%$\\
    $\text{EWC}_{\lambda:0.5k}$ &  $50.78\%$ & $63.99\%$ & $74.53\%$\\
    \midrule
    & \multicolumn{3}{c}{ImageNet (T-1) $\to$ Scenes (T-2)}\\
    \midrule
    FT  &  $53.56\%$ & $66.39\%$ & $74.70\%$\\
    LwF &  $65.27\%$ & $67.98\%$ & $74.78\%$\\
    $\text{EWC}_{\lambda:8k}$ &  $64.14\%$ & $68.50\%$ & $72.69\%$\\
    $\text{EWC}_{\lambda:0.5k}$ &  $54.61\%$ & $65.94\%$ & $74.70\%$\\
  \bottomrule
  \end{tabular}
  \caption{Forgetting of Task 1 measured via optimal linear probes (LP). Note that although the forgetting is much higher for finetuning compared to LwF, the LP accuracy is nearly identical, especially for the ImageNet $\to$ Scenes task, suggesting that LwF does not improve over naive finetuning in terms of forgetting knowledge acquired on ImageNet.}
  \label{tab:ImageNet-based-studies}
}
\end{table}
Although we followed the training procedure as closely as possible to the ones reported by~\cite{li2017learning}, the results are slightly different from the ones reported in~\cite{li2017learning} due to (a) not using the task-head warm-up step, where the backbone network is first frozen and the newly added task head is trained until convergence (warm-up), and then the entire network is trained until convergence, and (b) stochasticity of training neural networks. \Cref{tab:lwf-vs-ours} highlights these differences.

\begin{figure*}[!hb]
\vspace{-10pt}
\CenterFloatBoxes
\begin{floatrow}
\ffigbox[0.52\textwidth]{%
    \includegraphics[width=\linewidth]{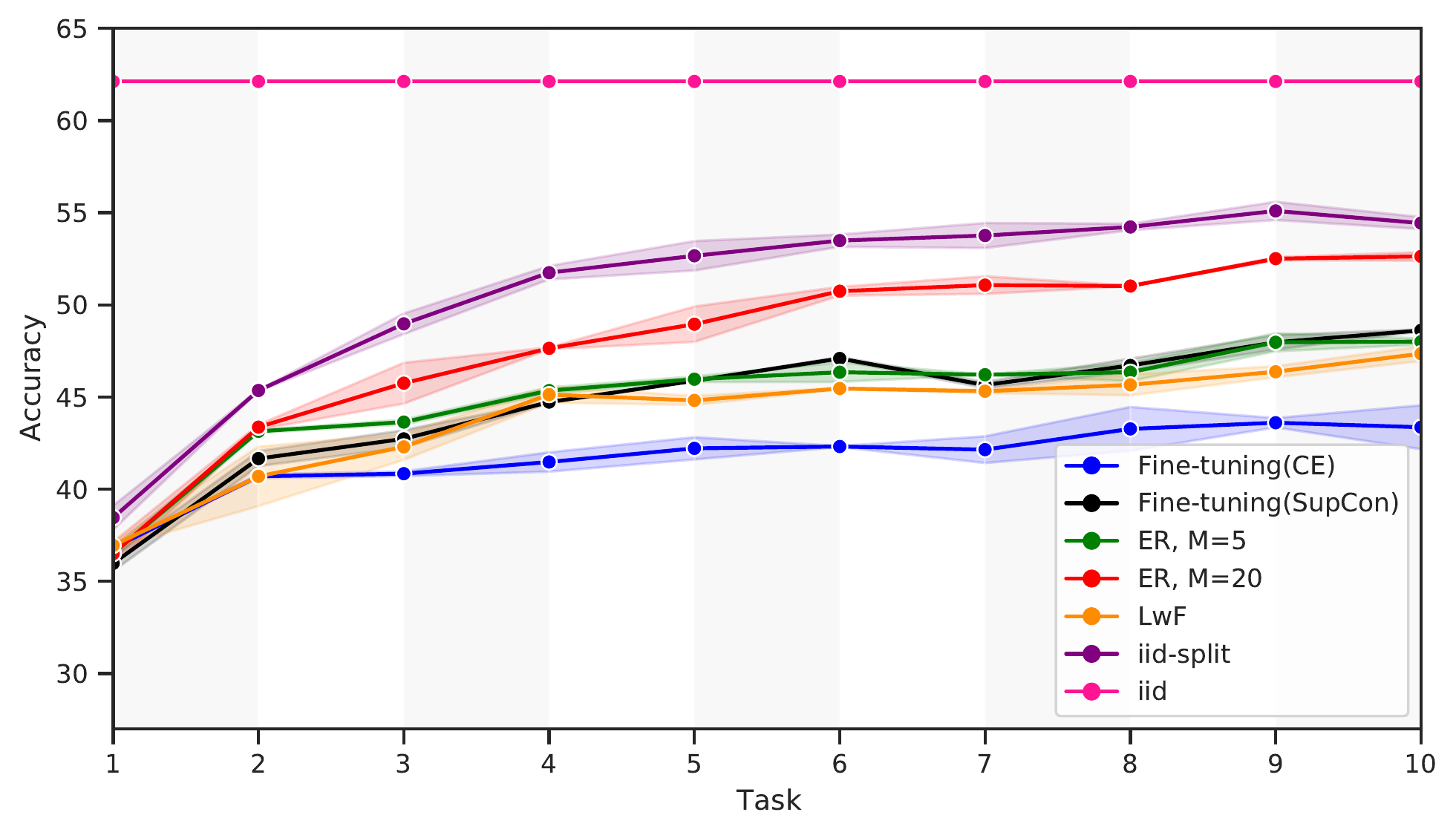}
}{%
  \caption{\small All-LP anytime evaluation plot on SplitCIFAR100 10-task sequence. All-LP is a probe trained on all, \textit{i.e.} seen and unseen tasks, training data and evaluated on all test data. We compare this with splitting iid data into 10 subsets trained in sequence, denoted as iid-split.
  \label{fig:cifar100_iid_split}}%
}
\capttabbox{%
\small
  \begin{tabular}{c|c|cc}
  \multicolumn{2}{c}{ }                         &Task 1 LP  & LP Acc. All \\
\multicolumn{1}{c}{}    & \multicolumn{1}{c}{}& T=10  & T=10\\\hline
                  & RN18, Width=32  &   70.4  & 74.0  \\  
fine(SupCon)  & RN18, Width=128 &   74.3  & 77.1  \\  
                  & RN101, Width=32 &   71.9  & 75.4  \\  \hline
                  & RN18, Width=32  &   64.8  & 70.8  \\  
fine(CE)      & RN18, Width=128 &   70.5  & 74.2  \\  
                  & RN101, Width=32 &   67.9  & 72.4  \\  \hline
                  & RN18, Width=32  &   74.2  & 75.7  \\  
ER-M5             & RN18, Width=128 &   75.6  & 77.3  \\  
                  & RN101, Width=32 &     74.5  & 76.1  \\  \hline
                  & RN18, Width=32  &   76      & 76.4  \\  
ER-M20              & RN18, Width=128 &   78.8  & 80.1  \\  
                  & RN101, Width=32 &   77.1  & 77.5  \\  \hline
                  & RN18, Width=32  &   70.1  & 73.4  \\  
LwF                 & RN18, Width=128 &   74.8  & 76.7  \\  
                  & RN101, Width=32 &   71.0  & 74.6  \\  

  \end{tabular}
}{%
  \caption{\small Final Accuracy of 10 task SplitCIFAR100 sequence with variable width and depth in the offline setting. $M$ indicates the number of samples per task used in the ER buffer. We compare SupCon LP to others showing it has similar improvements in width and depth to non-finetuning methods.
  \label{tab:CIFARLongTask_Offline_supp}}%
}
\end{floatrow}
\end{figure*}
\begin{figure}[b]
    \centering
    \includegraphics[width=0.49\textwidth]{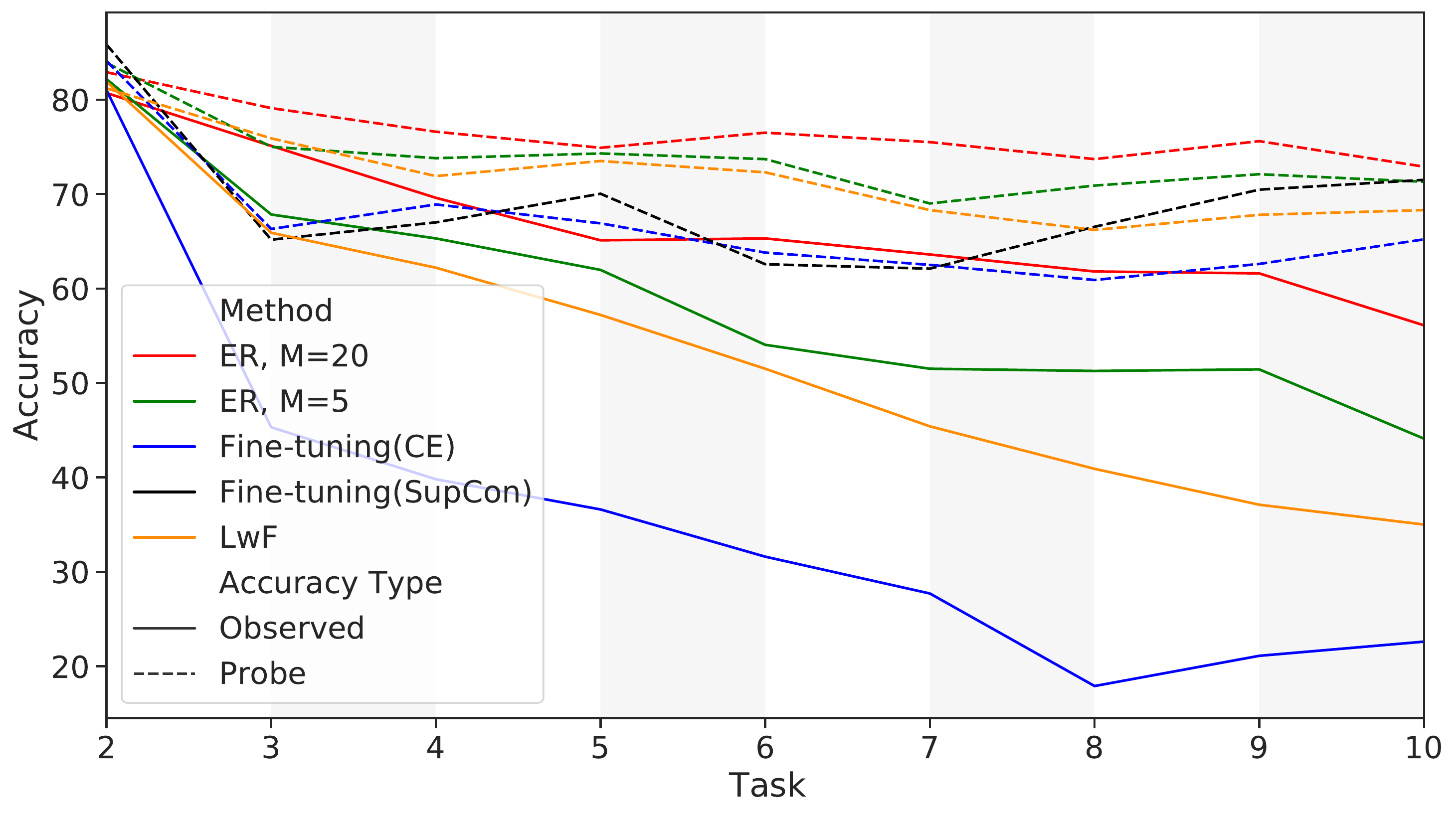}\label{fig:cifar100_task2_supp}
    \includegraphics[width=0.49\textwidth]{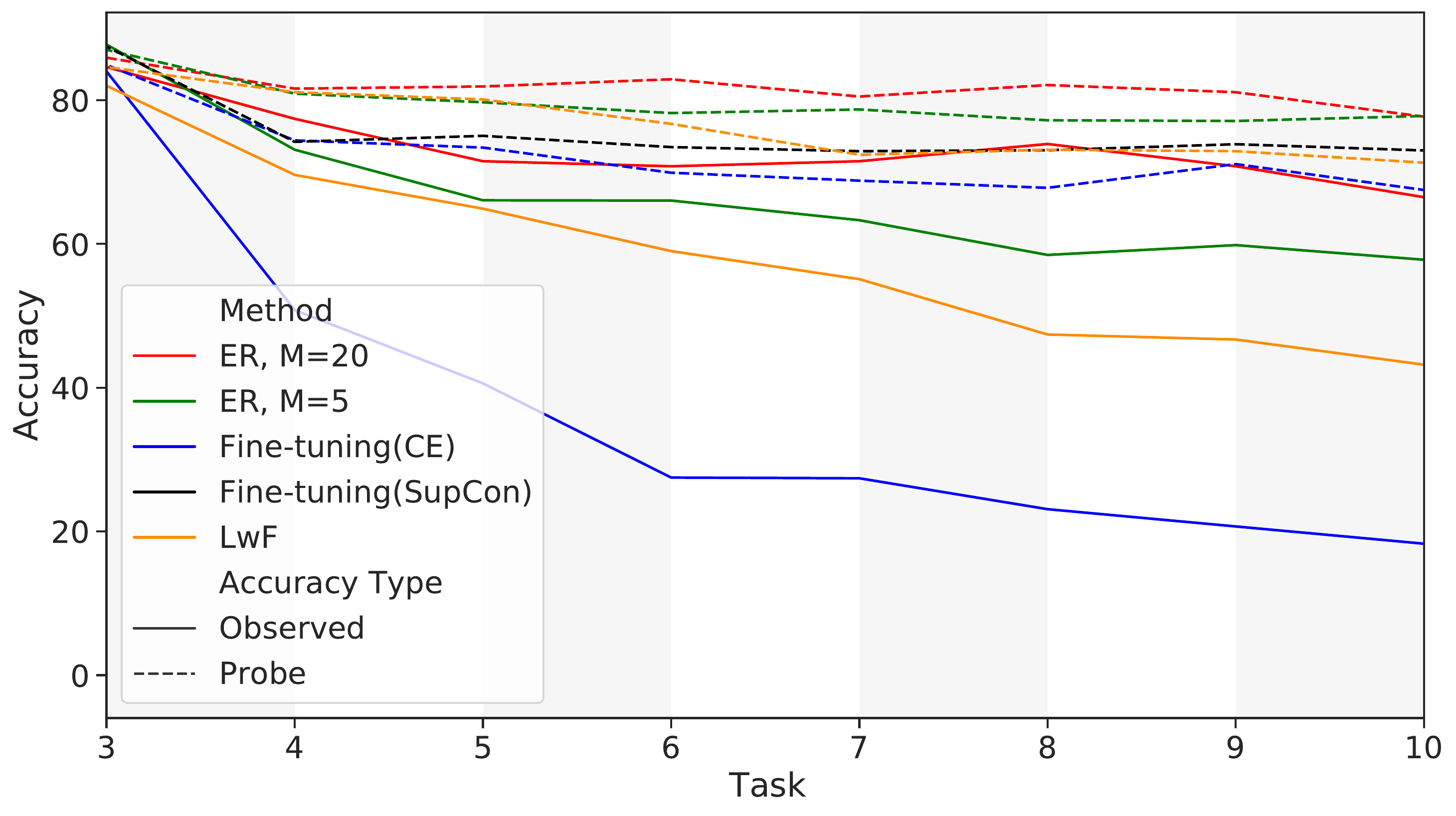}\label{fig:cifar100_task3_supp}\\\centering
    \includegraphics[width=0.49\textwidth]{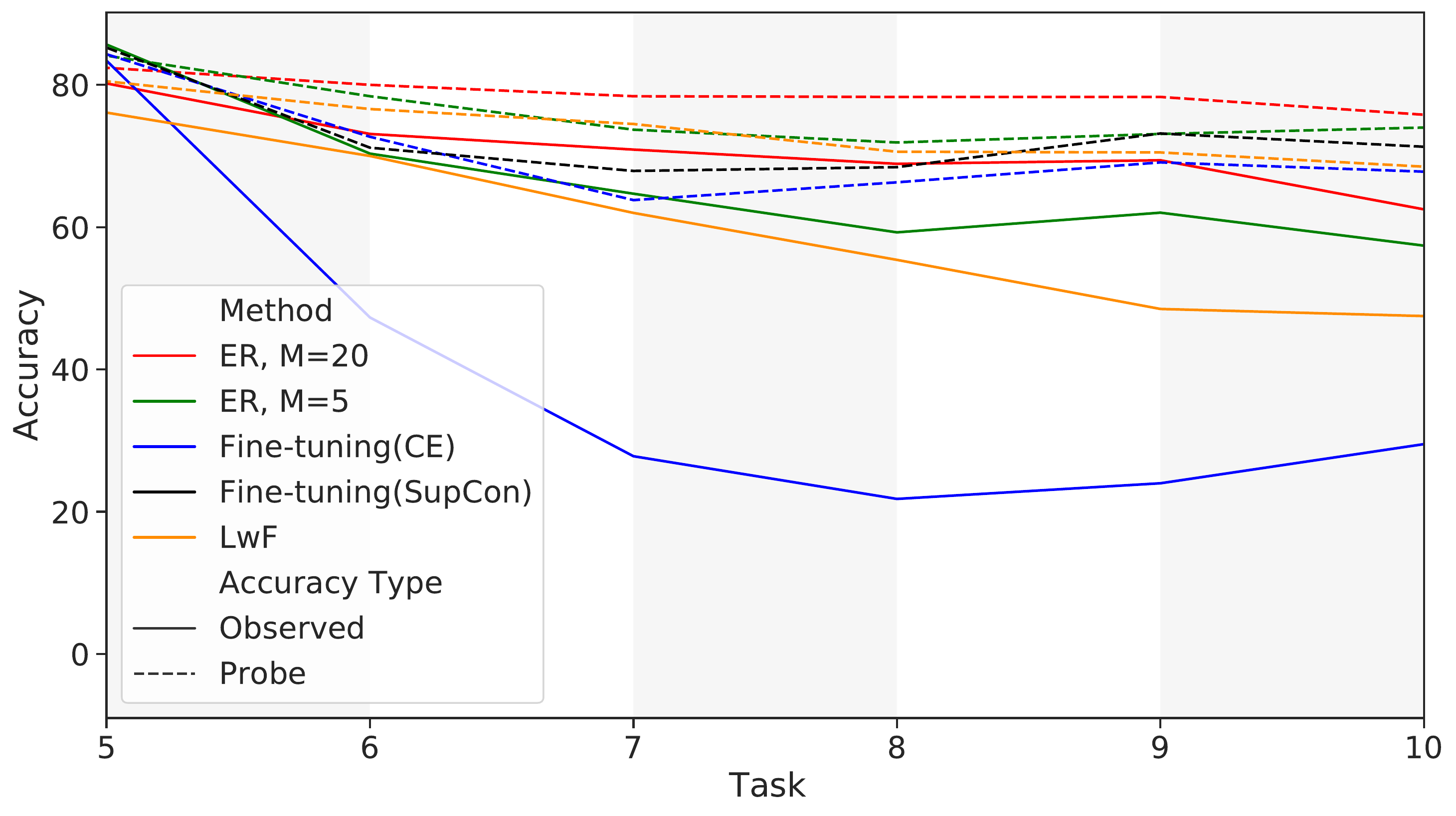}\label{fig:cifar100_task5_supp}
    \caption{LP and Observed accuracy for Task 2 (upper left), 3 (upper right), and 5(bottom)  on 10-Task SplitCIFAR100.}
    \label{fig:cifar100_taskall_supp}
\end{figure}
\vspace{-1pt}
\subsection{Comparing Overall Representation Improvement}
\vspace{-1pt}
In addition to representation forgetting, we consider also measuring how much a representation improves overall as seen by a linear probe trained and evaluated on data from the union of all current and future tasks. We can evaluate this by training at each step in the sequence an ``All-LP", that is a linear probe trained on all the training data and evaluated on all test data. A natural baseline to compare such an approach to is splitting iid data into 10 subsets trained in sequence (we denote this iid-split). The results of this evaluation are shown for SplitCIFAR100 in Figure~\ref{fig:cifar100_iid_split}. We observe again that SupCon exceeds LwF methods and competes with the small replay-sized ER method.

\vspace{-1pt}
\subsection{Representation Forgetting for Other Tasks}
\vspace{-1pt}
Complementing the results in \cref{sec:observed-vs-lp-acc}, we report the observed and LP accuracies for methods when measured for tasks beyond Task 1. Specifically for SplitCIFAR100 we show trends for Task 2, 3, and 5 in~\cref{fig:cifar100_taskall_supp}. We observe similar trends as reported for Task 1 in~\cref{sec:observed-vs-lp-acc}.

\vspace{-1pt}
\subsection{Effect of Increased Model Capacity for SupCon}
\vspace{-1pt}
To complement the results in~\cref{sec:effect-of-increased-model-capacity}, we also compare finetuning with SupCon and CE in terms of its properties on wide and deep networks. Since SupCon training does not have an observed accuracy we compare only the LP performance unlike~\cref{sec:effect-of-increased-model-capacity}. From~\cref{tab:CIFARLongTask_Offline_supp}, we observe that similar to CE the LP performance improves with depth and exceeds that of LwF~\cite{li2017learning}, and nearly matching ER-M5 with greater width.

\end{document}